# A Systematic Review on Sleep Stage Classification and Sleep Disorder Detection Using Artificial Intelligence

Tayab Uddin Wara[†], Ababil Hossain Fahad[†], Adri Shankar Das[†], Md. Mehedi Hasan Shawon[†*]

*tayabuddinwara@gmail.com*, *ababilhossainfahad@gmail.com*, *adrishankar@gmail.com*, *shawon.cse.bracu@gmail.com*

*BSRM School of Engineering, BRAC University, Dhaka, Bangladesh.*

## Abstract

Sleep is vital for our physical and mental health, and sound sleep can help us to focus on daily activities. Therefore, a sleep study that includes sleep patterns and disorders is crucial to enhancing our knowledge about individual health status. The findings on sleep stages and sleep disorders relied on Polysomnography (PSG) and self-report measures, and then the study went through clinical assessments by expert physicians. Artificial Intelligence (AI) has made the evaluation process of sleep stages and sleep disorders classification more efficient. Many studies have focused on analyzing various datasets using advanced techniques and algorithms to improve computational ease and accuracy. This review aims to comprehensively analyze recent literature on different approaches and outcomes in sleep studies, specifically on 'sleep stages classification' and 'sleep disorder detection' using AI. Initially, 185 articles were selected from top journals, and eventually, 81 of them were reviewed in detail, covering the period from 2016 to 2023. Brain waves are the most commonly used body signals for studying sleep patterns and disorders. Almost 36% of the research exclusively used brain activity signals, and 80% combined them with other body parameters in sleep staging. The Neural Network (NN) algorithms are the most popular, having 47% of the total usage. At the same time, Long Short-Term Memory (LSTM), Ensemble Learning (RL), Support Vector Machine (SVM), and Random Forest (RF) accounted for 15%, 12%, 7%, and 6% of usage, respectively. For evaluating AI model performance, accuracy or precision is used in 86.42% of cases, followed by an F1 score of 46.91%, Kappa of 39.51%, Specificity of 30.86%, Sensitivity of 29.63%, along with other metrics. This article would help physicians and researchers get the gist of AI's contribution to sleep studies and the feasibility of their intended work.

## 1. Introduction

Sleep is considered one of the most important aspects of a healthy physical and mental state. Body functioning tends to be disoriented if any hindrance occurs in regular sleep. For example, a recent study showed that sleep quality affects our regular academic performance [1]. Furthermore, according to the National Sleep Foundation survey, disturbed sleep patterns have been found among 40% of patients with different diseases and disorders [2]. As a result, sleep study analysis and monitoring are essential for overall quality and a healthy life. Sleep is mainly categorized into five different stages according to the American Academy of Sleep Medicine (AASM): Wake (W), Rapid Eye Movement (REM), and Non-rapid Eye Movement (NREM), where NREM is further divided into three stages: N1, N2, and N3. The transitional stage between wakefulness and sleep is called stage N1, or light sleep. The deeper stage of light sleep, when the body begins to relax further, is called the stage N2, or intermediate sleep. The deepest stage of NREM sleep is called stage N3 or delta sleep. The REM sleep stage is characterized when a person is deeply asleep but continuously moves their eyes. Although sleep is a natural process of calmness of voluntary and sensory muscles, our brain constantly works during sleep. The Electroencephalogram (EEG) signals collected from the brain show variable patterns and aperiodic signals and can rapidly change over several hours during sleep. All these waves of EEG signals are interpreted in 30-second frames to categorize and identify the sleep stages [3]. Sleep studies provide information about sleep stages and disorders. Some common sleep disorders may include Insomnia, REM Sleep Without Atonia (RSWA), Obstructive Sleep Apnea (OSA), Idiopathic REM sleep behavior disorder (iRBD), and so on. Due to their prevalence and harmful physiological effects, these disorders affect a significant portion of the population [4,5,6]. Insomnia is a common sleep disorder characterized by persistent difficulties in falling asleep or experiencing non-restorative sleep despite enough opportunity, which leads to long-term health consequences and impaired daytime functioning. OSA involves frequent episodes of airway obstruction during sleep, which results in intermittent breathing pauses, fragmented sleep, and decreased oxygen levels. This condition is related to an increased risk of cardiovascular diseases, cognitive impairment, and metabolic disorders. iRBD is characterized by the loss of normal muscle atonia during the REM stage of sleep, which allows individuals to act out vivid dreams with the potential for self-injury. iRBD may be an early indicator of neurodegenerative diseases such as Parkinson's disease and other synucleinopathies. Each disorder manifests through distinctive patterns in sleep signals that differ from those of healthy individuals. These abnormalities in sleep patterns are identified through a detailed analysis of PSG data, allowing an expert physician to diagnose specific disorders based on the degree and nature of the deviations observed.

*Corresponding author.
† Equal Contribution in Manuscript Preparation.

We observed numerous works on sleep studies where AI plays a vital role in sleep stages and sleep disorders diagnosis more efficiently. For instance, SleepEEGNet is a deep learning approach that achieved an overall accuracy of 84.26% in sleep stage scoring [7]. Similarly, another work demonstrated the effectiveness of transfer learning in wearable sleep stage classification using photoplethysmography [8], and other authors used a wrist-worn device to collect the relevant accelerometer and applied an RF algorithm to predict sleep classification [9]. Several ML algorithms, such as RF, K-Nearest Neighbours (KNN), SVM, and a hybrid neural network model, were utilized that gained 87.4% accuracy in the sleep stage classification [10]. Another study introduced a model called U-Sleep, publicly available for automated sleep staging and based on a deep-learning architecture [11]. Additionally, 'Oura Ring' was introduced to collect multiple sensor data like accelerometer, autonomic nervous system (ANS)-mediated peripheral signals, and circadian, then applied a Machine Learning (ML) model to sleep stage detection with a maximum accuracy of 94% [12].

Moreover, numerous examples of sleep disorder detection using AI models are also observed. For instance, an explainable Convolutional Neural Network (CNN) was introduced to detect sleep apnea from single-channel EEG recordings [13]. The network learned frequency-band information associated with known sleep apnea biomarkers and showed good generalization across subjects with an accuracy of 69.9% and a Matthews correlation coefficient of 0.38. Another paper presents a sleep-monitoring model for detecting OSA using a CNN-LSTM architecture on single-channel Electrocardiogram (ECG) signals, segmented with a 10-second sliding window, which achieved 96.1% accuracy, 96.1% sensitivity, and 96.2% specificity on the Apnea-ECG dataset, outperforming baseline methods [5]. Furthermore, a different study utilized the Stanford STAGES dataset and proposed an interpretable ML solution for screening major depressive disorder using 5-minute nighttime ECG signals [14]. The Bayesian-optimized extremely randomized trees classifier had the best accuracy at 86.32%, highlighting the importance of gender in the model's prediction and suggesting potential integration into portable ECG monitoring systems.

This systematic literature review (SLR) is motivated by the absence of a comprehensive review of the latest advancements in using AI for sleep stage classification and sleep disorder detection. While a systematic literature review on sleep apnea detection using deep learning was published in early 2023, it needs to cover predicting other sleep disorders using AI [15]. On the other hand, there is a review article on automatic sleep stage classification, which lacks information on sleep disorders using AI [16]. Therefore, a complete review is necessary to understand the overall impact of AI applications in sleep study analysis and to explore recent findings. This review paper encompasses recent articles on sleep stages and disorders using AI, showcasing various methodologies and techniques applied to sleep studies. Our key contributions to this article are:

- We have analyzed the recent findings of 81 articles on sleep stage classification and sleep disorder detection from 2016 to 2023 using AI. The distribution of articles by year is illustrated in Figure 1.
- We have identified the most commonly used AI models, body parameters, sample data size, and model evaluation metrics for sleep stage classification and sleep disorder detection.

In this paper, we have organized our investigation into several sections to provide a comprehensive outline of the use of AI in sleep study analysis. In section 2, we acknowledge the necessity of AI, highlighting the potential to reduce manual labor and operational costs. Section 3 discusses the methodology and details of our research steps and strategies, including formulating research questions and abstracting our search strategy to identify relevant literature. Section 4 discusses AI models and their working principles, illustrating their mechanisms through block diagrams. We also examine the advantages and disadvantages of these models to guide researchers in selecting the most suitable model for specific objectives. Subsequently, in section 5, we discuss the results in detail, where we analyze the 81 selected articles to identify commonly used body parameters, sample data size, AI models, and evaluation metrics in the context of sleep study analysis. In Section 6, we summarized the key findings from the selected articles. Finally, in section 7, we discuss the future research directions, pointing out study gaps and outlining the potential future scope of research in this field. This section draws insights from the reviewed publications and provides a roadmap for future studies.

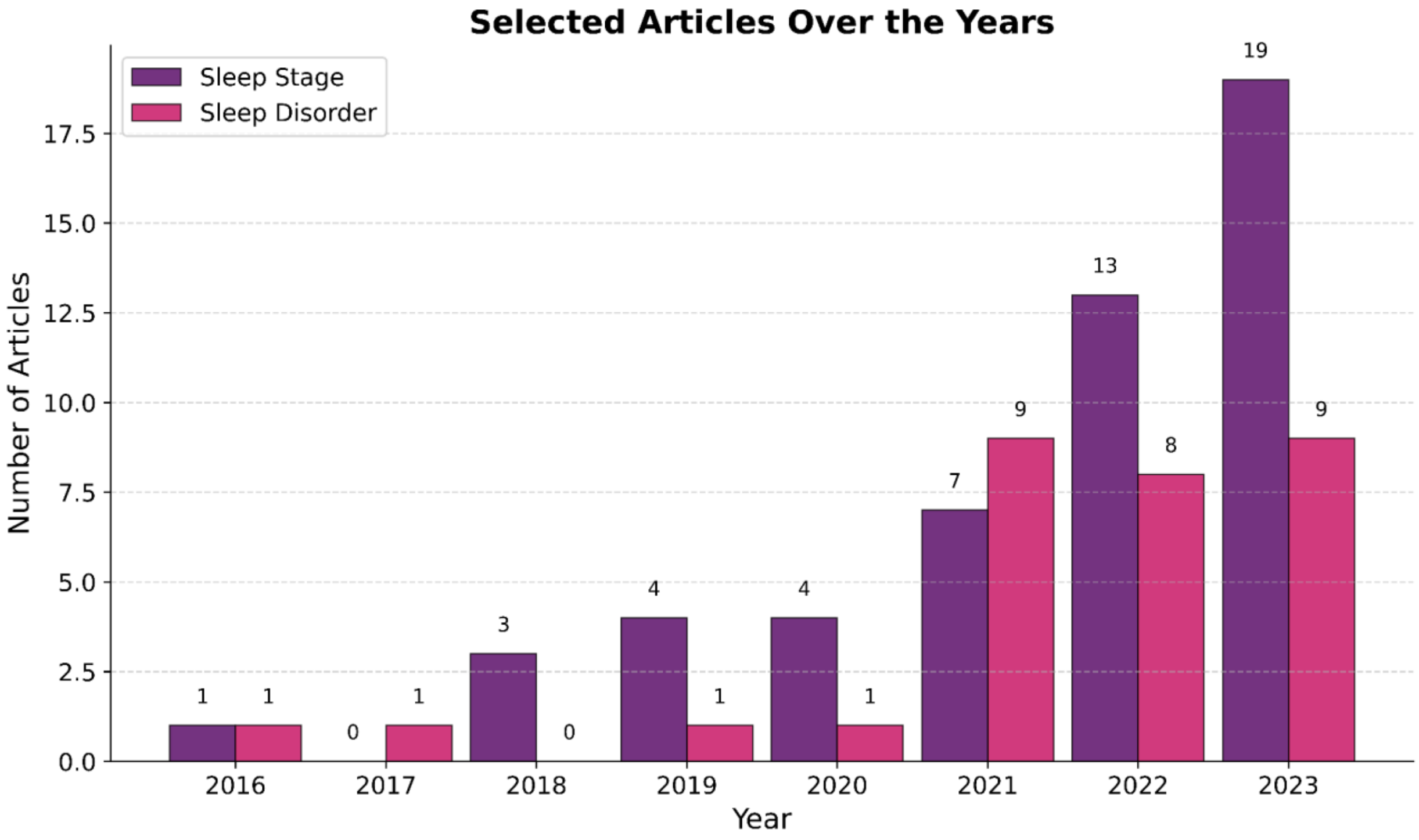

**Figure 1.** The number of articles considered over the last 8 years from 2016 to 2023 for this SLR.

## 2. Necessity of Artificial Intelligence in Sleep Studies

A sleep study involves observing patients throughout the night and capturing their PSG data. PSG includes multivariate biological signals such as Electrocardiograms (ECG) for heart activity, EEG for brain waves, Electrooculograms (EOG) for eye movement records, respiratory flow, and blood oxygen saturation level, and Electromyograms (EMG) for muscle movement. After the overnight sleep data is collected, an expert technician classifies the sleep stages based on the collected data and sends them for interpretation by doctors specializing in sleep to check for sleep disorders. After examining the sleep study methods, it is evident that the traditional manual approach is quite complex. Since many people are facing sleep-related disorders, it becomes essential to have a robust and consistent method to help the physician deal with the complexity and create a precise report system for diagnosing and treating such disorders. However, this process is labor-intensive, time-consuming to prepare the report, and has a high operational cost. Moreover, this process is also prone to human-made errors in correctly classifying sleep stages. AI algorithms, including ML and neural network models, have shown promising results in different research fields, including biomedical research. As a result, sleep study analysis has undergone a significant transformation with the widespread adoption of AI for predicting sleep stages and identifying disorders. This transformation is evident in the increasing number of studies utilizing AI models for sleep stage classification and disorder detection, which enables researchers and healthcare professionals to gain deeper insights into sleep patterns and disorders. AI has facilitated and automated this sorting by feeding the signals or after pre-processing signals into the AI models, which replace the laborious manual comparison process to identify the sleep stages accurately. Thus, valuable time and resources are saved, and healthcare facilities can operate and allocate resources more efficiently. Outstanding computational advances with sophisticated algorithms have improved the efficiency of diagnosing and screening disorders. For instance, analyzing the sleep stages, narcolepsy [17], and prediction or estimation of OSA have been seeing promising results [18]. Moreover, timely assessments of sleep-related issues can be done by replacing intensive manual labor and repetitive tasks, which will benefit healthcare providers by streamlining their workflow, saving patients money, and ensuring quality treatment on time.

## 3. Methodology

A review paper provides an overview of progress in a particular field or topic depending on the currently available literature so that a reader gets a brief idea of that specific field by reading only one paper, there are different types of review papers, such as narrative and systematic. We have accomplished this paper as an SLR since we gathered information from empirical research to answer specific research questions and analyze and present the findings from the selected articles. In this paper, we have reviewed the sleep study, which includes sleep stage classification and sleep disorder detection analysis that is being accomplished using AI. The work method can be divided into three phases. At first, in phase 1, different online journals were utilized to search for relevant works using specific keywords described in section 3.2.1. The papers beyond the scope of this research were removed. We fixed some criteria to exclude those papers, and as a result, we finally got 81 papers for analysis, as shown in Figure 2. After gathering relevant literature, several research questions were set to analyze the literature, described in section 3.1. Different parameters were investigated, such as the best-fitted model, performance, data source, sample size of the data sets, the signal utilized for the study, and type of study, whether it is sleep stage classification or sleep disorder detection. In phase 2, all 81 papers were thoroughly reviewed, and all the mentioned parameters were extracted. This phase reflects the collaborative work as 81 papers were distributed among the authors to read and collect the data. This parallel working technique significantly reduced time and effort, making the data collection smooth and accurate. In phase 3, collected parameters were analyzed, and recent trends, frequency distributions, and work variation can be observed using visual illustration in section 5. By examining the trends and graphs, the research questions were addressed. Finally, future scopes are determined by analyzing all the papers and this study.

### 3. 1 Research Questions

**Research Question 1. What different body parameters were used in sleep studies for AI applications?**

During the overnight monitoring, physiological signals, such as EEG, EOG, EMG, and ECG, are collected, along with breathing functions, respiratory airflow, peripheral pulse oximetry, etc. However, we want to determine the most commonly used body parameters for sleep studies in order to apply AI models and obtain the best possible results. We would also like to investigate the different body-generated signals and dependencies for which performance fluctuates.

**Research Question 2. What sample dataset size was considered in the experiments?**
The types and numbers of datasets are crucial factors in training AI models, impacting their performance, efficiency, and accuracy. This question investigates the significance of the diverse samples included in different researchers' work with frequency usage, and the findings will answer this question.

**Research Question 3. Which AI algorithms were best fitted in the experiment?**
To improve the classification of sleep stages and detection of sleep disorders, it is essential to use widely used, advanced AI algorithms. This will help reduce the time and effort spent exploring and using different models to achieve the best results.

**Research Question 4. What were the evaluation methods and performance metrics of the experiment?**
It is important to use various evaluation methods, such as validation and performance metrics, to determine the robustness of AI models. Different validation techniques help measure performance, including accuracy, F1 score, kappa, sensitivity, and specificity. Quantitative analysis of the model's performance and evaluation is crucial for accurately predicting the expected outcomes.

## 3.2 Search Strategy

Table 1 shows our paper selection strategy from different journal databases, and Figure 2 shows the number of papers for initial selection, the criteria for which papers were discarded automatically and manually, and the number of papers left for the final review.

### 3.2.1 Keyword Selection

We set four specific keywords to search our relevant papers, and these search items were mainly applied to the title and abstract. These are sleep studies with artificial intelligence, sleep studies with machine learning, sleep stage classification, and sleep disorder and/or detection. Different authors may have used different synonyms for these keywords, but they worked within the scope of our study. Therefore, we chose keywords to represent all the works in this review article.

**Table 1: Keyword Search**

| Database | Search query used |
|---|---|
| Springer, Nature, PLOS One, PubMed, Science Direct, MDPI, Hindawi, Peerj | ("sleep study with artificial intelligence" OR "sleep study with AI") AND ("sleep study with machine learning" OR "sleep study with ML") AND ("sleep stage" OR "sleep stage classification" OR "sleep classification") AND ("sleep disorder and detection" OR "sleep disorder" OR "sleep detection") |

### 3.2.2 Screening

From the different sources mentioned in the PRISMA diagram, we initially selected 185 publications. After that, we removed 60 papers because they were duplicates or inaccessible. Moreover, we removed 16 other publications because the papers did not use AI models and were outside the range from 2016 to 2023. Furthermore, 28 articles were removed because they didn't match the subject of this study. After all the steps, we finally left 81 articles to study in this paper, including 29 sleep disorders and 52 sleep-stage classification articles.

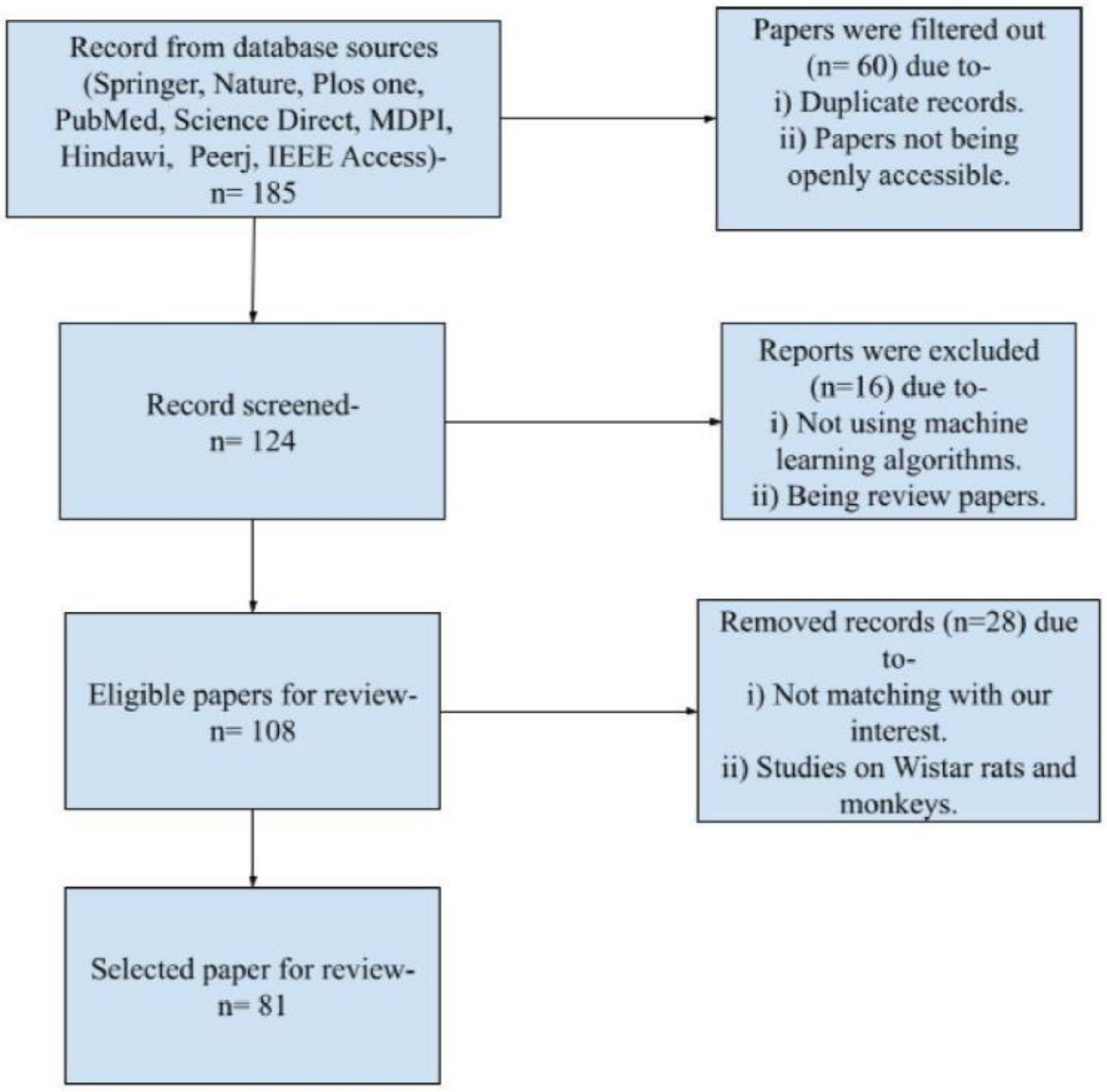

**Figure 2.** PRISMA diagram shows how the articles were selected for review.

## 4. Artificial Intelligence Models

While reviewing the papers, we discovered 32 unique AI models. We have detailed the five most frequently used models' structure, features, advantages, and disadvantages.

### 4.1. Convolutional Neural Network

CNN is an artificial neural network commonly used for signal or image classification. It consists of two main sections: feature extraction and classification. In the feature extraction section, input, convolution, and pooling occur. In the classification section, there is the fully connected layer and also the output layer. In the input layer, all the raw data is fed. The convolutional layer uses convolutional operations to detect edges and textures. This layer extracts important features from the data. The pooling layer samples the feature maps to reduce spatial dimensions. All the neurons from the previous layers and subsequent layers are connected in the fully connected layer. Then, the output layer produces the final output, as shown in Figure 3 [19].

#### 4.1.1 Advantages of CNN

CNN is widely known for its efficient recognition algorithm. The CNN architecture is straightforward and needs fewer training parameters. This model's noise sensitivity is reduced because of its pooling layers and feature learning. As there are no model parameters to train in the pooling layers and the number of model parameters is comparatively low, the algorithm is fast. This feature of learning helps it learn simple patterns to complex structures automatically. It uses a parameter-sharing feature to reduce the number of parameters for efficiency. CNN can parallelize its operations to take advantage of modern GPU architectures and speed up the process.

#### 4.1.2 Disadvantages of CNN

The CNN model needs significant labeled data points for efficient training. With sufficient data, it may recognize new or unseen examples. It requires very high computational power as it uses several layers and parameters. Another drawback of the model is that it needs a fixed-size input, making it less flexible. Another issue could be if the training data becomes biased for any reason. CNN will start learning that biased information, and the model will provide biased output while applied in modern-day applications.

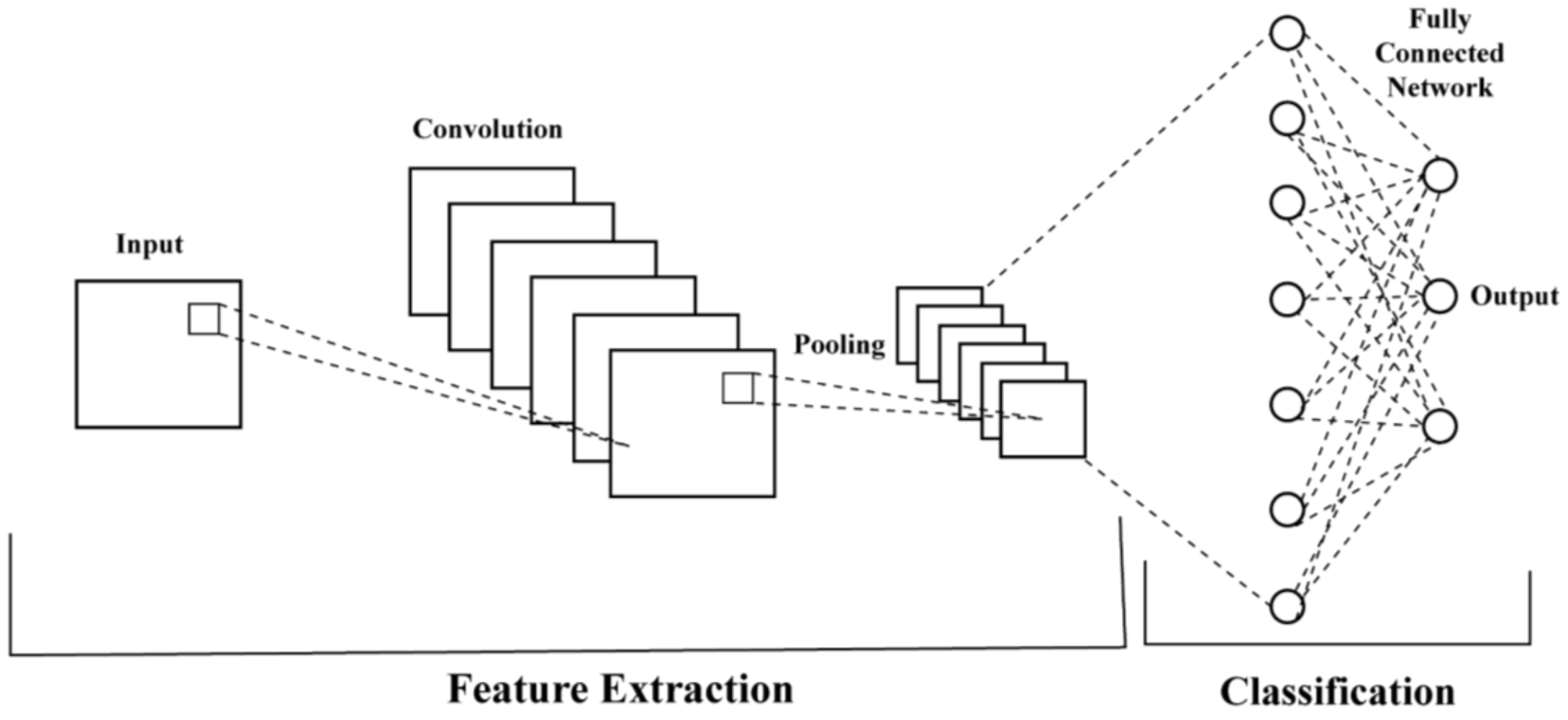

**Figure 3.** CNN block diagram.

### 4.2 Recurrent Neural Network

A Recurrent Neural Network (RNN) is a neural network used to process sequential data. The RNN models use the supervised learning concept. They contain feedback connections used to store information from earlier stages [20]. This feature makes them more convenient for tasks requiring temporal dependencies or context, such as time series analysis, speech recognition, and natural language processing. The RNN unfolding architecture is presented in Figure 4.

#### 4.2.1 Advantages of RNN

The main advantage of the model is that it can take and process inputs of any size. Moreover, the information-retaining feature of this model is very convenient when used for any time series prediction. Another advantage is that the input size does not increase the model size. RNN also has a parameter-sharing feature. It can share the weights across the time steps.

#### 4.2.2 Disadvantages of RNN

RNNs sometimes can suffer from gradient vanishing or exploding problems during training while processing long sequences. This drawback can hamper long-term dependency learning. Training RNNs is also challenging and computationally costly. The sequential nature of the model slows down the training process. As the model retains past information, it needs much storage. Moreover, RNN is more sensitive than other neural networks in the case of hyperparameter choices. These models are also less suitable for irregular time intervals.

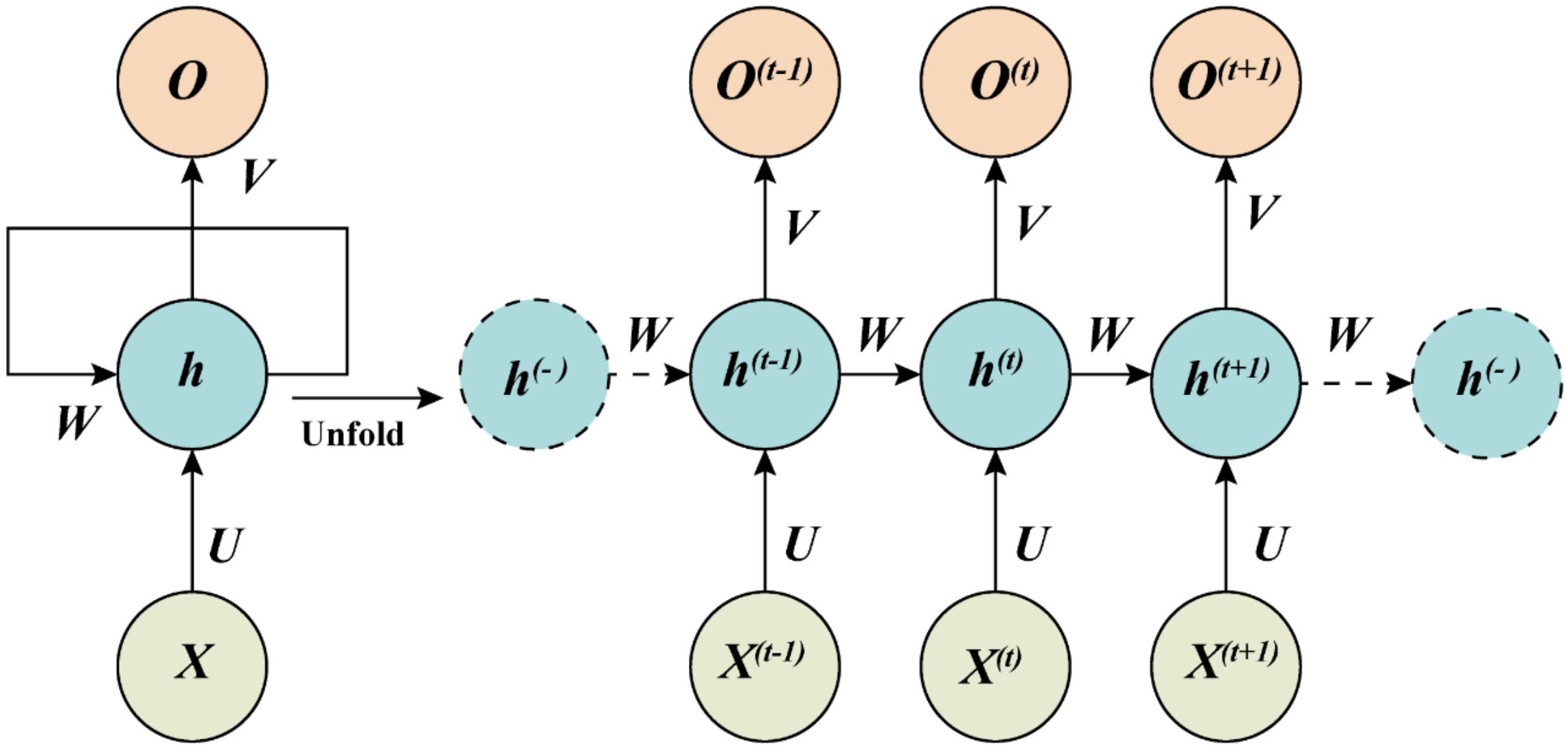

**Figure 4.** RNN block diagram.

### 4.3 Support Vector Machine

SVM algorithms are used to perform classification and regression tasks. They are used when clear margins among different classes are crucial. As shown in Figure 5, the SVM determines a hyperplane that determines the best fit between the classes. The vectors of particular classes generate hyperplanes, which is why the model is called support vectors. This model aims to maximize the optimal hyperplane and ensure a reliable prediction [21].

### 4.3.1 Advantages of SVM

The SVM can perform both classification and regression. It uses kernel functions to manage linear and nonlinear decision boundaries, making the model versatile. The model can also perform efficiently with numerous features, making it practical for high-dimensional datasets. Moreover, it is durable enough to overfit when the feature number exceeds the sample, making it suitable for small datasets. Besides, SVM provides high-quality results.

### 4.3.2 Disadvantages of SVM

SVM also has some disadvantages. SVM is very sensitive to noise. Moreover, SVM is computationally intensive while training, making it less efficient for large datasets. Besides, the model is comparatively slower for nonlinear data than other models. SVMs can also be memory intensive while dealing with large datasets, which limits their scalability and increases runtime.

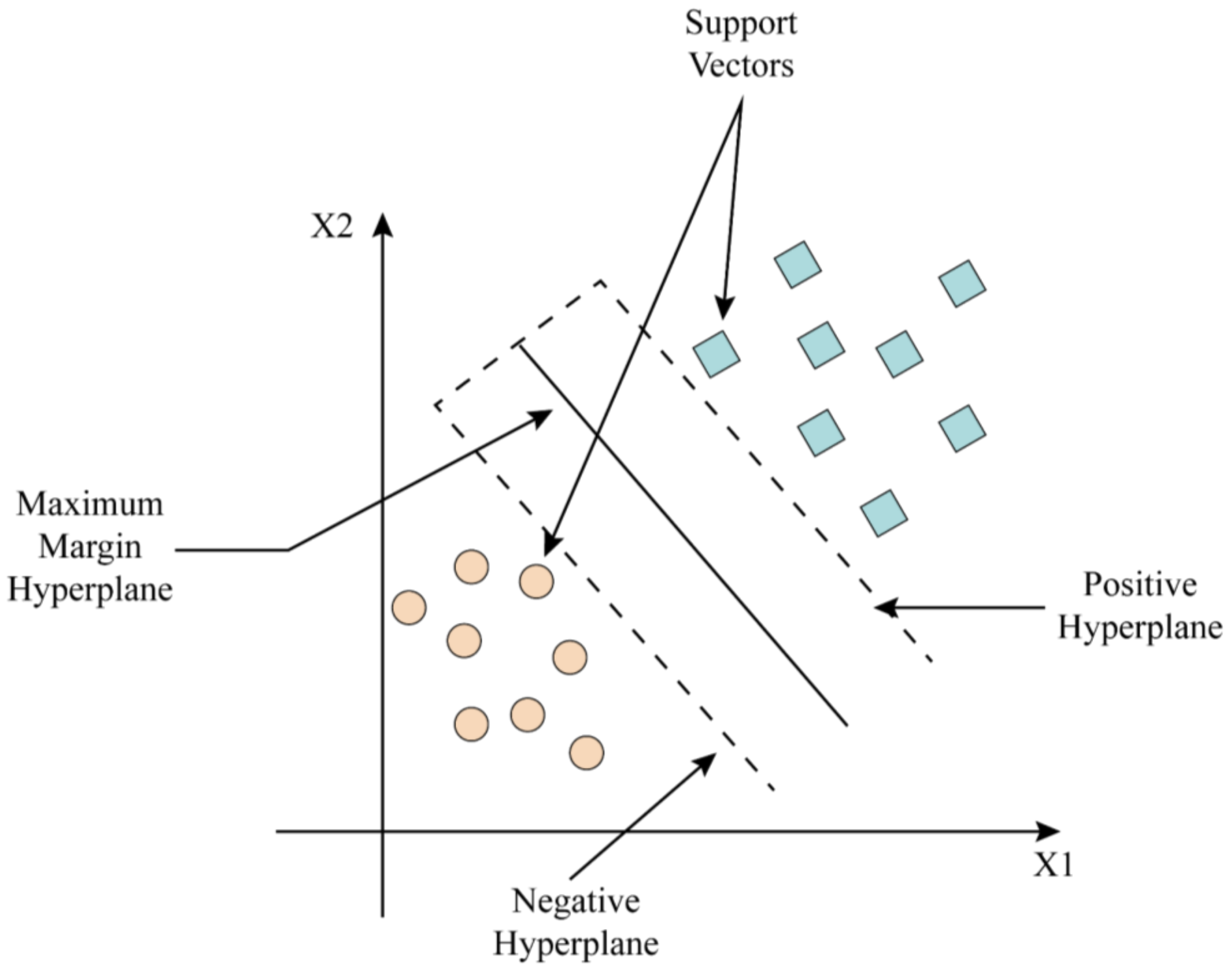

**Figure 5.** SVM block diagram.

### 4.4 Long Short-Term Memory

LSTM is a form of recurrent neural network architecture mainly designed to beat the issues of learning long-term dependencies in sequential data. The LSTM can retain information longer than the RNN using its memory cell. The memory block of this model is small but can maintain memory for a longer time. The model uses three gates to control the memory block: the forget gate, the input gate, and the output gate, as shown in Figure 6. The forget gate decides which information will be stored and processed further. Any information is not required to be sent to the input gate. Only the needed information is sent to the input gate. The information is introduced and combined with the cell state in the input gate. Besides, in the output gate, the data from the input gate is multiplied by the cell's $tanh$ generated vector to get the output [22].

### 4.4.1 Advantages of LSTM

LSTM can retain information for a longer time. The model's forget gate can filter out unnecessary data. The model can be parallelized more efficiently than regular RNN. Moreover, LSTM can predict considerably faster and more accurately because of the long-time information retention feature.

### 4.4.2 Disadvantages of LSTM

LSTM can be computationally intensive compared to simple feedforward neural networks. Gradually, the model turns to get overfitted for a more notable amount of data. It also has weight initialization problems. Moreover, it involves several hyperparameters; therefore, proper tuning is necessary, and finding the perfect configuration may be difficult.

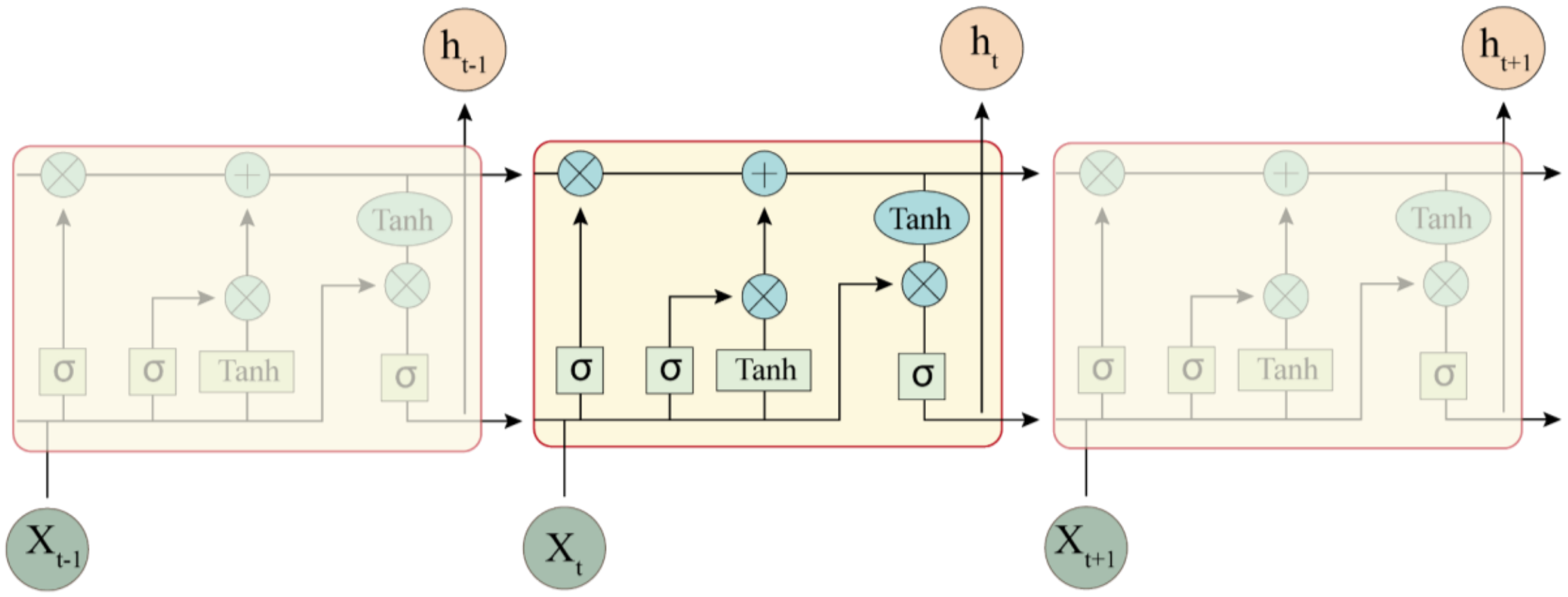

**Figure 6.** LSTM block diagram.

## 4.5 Random Forest

RF is a widely used machine learning algorithm for classification, regression, and other tasks. The model provides a single result by combining the output from multiple decision trees, as illustrated in Figure 7. The model is like a tree, where numerous trees spread from a single node. The trees also have multiple child nodes. More nodes spread from the previous child nodes, and the cycle continues several times. Therefore, it generates various results and combines them to get a single result [23].

### 4.5.1 Advantages of RF

RF is known for its accuracy, robustness, and ability to handle diverse tasks. The model provides high accuracy using the ensemble of its diverse decision trees. Moreover, the ensemble model helps create strong models that are less likely to be overfitting. The model is designed to handle missing data from the dataset without any imputation.

### 4.5.2 Disadvantages of RF

The main disadvantage of this model is that it can be very complex and difficult to interpret compared to the individual decision trees. It also has a disadvantage in terms of computational intensity. Moreover, more extrapolation is needed with this model. It may only extrapolate within the limited range of training data, limiting its performance on unseen data.

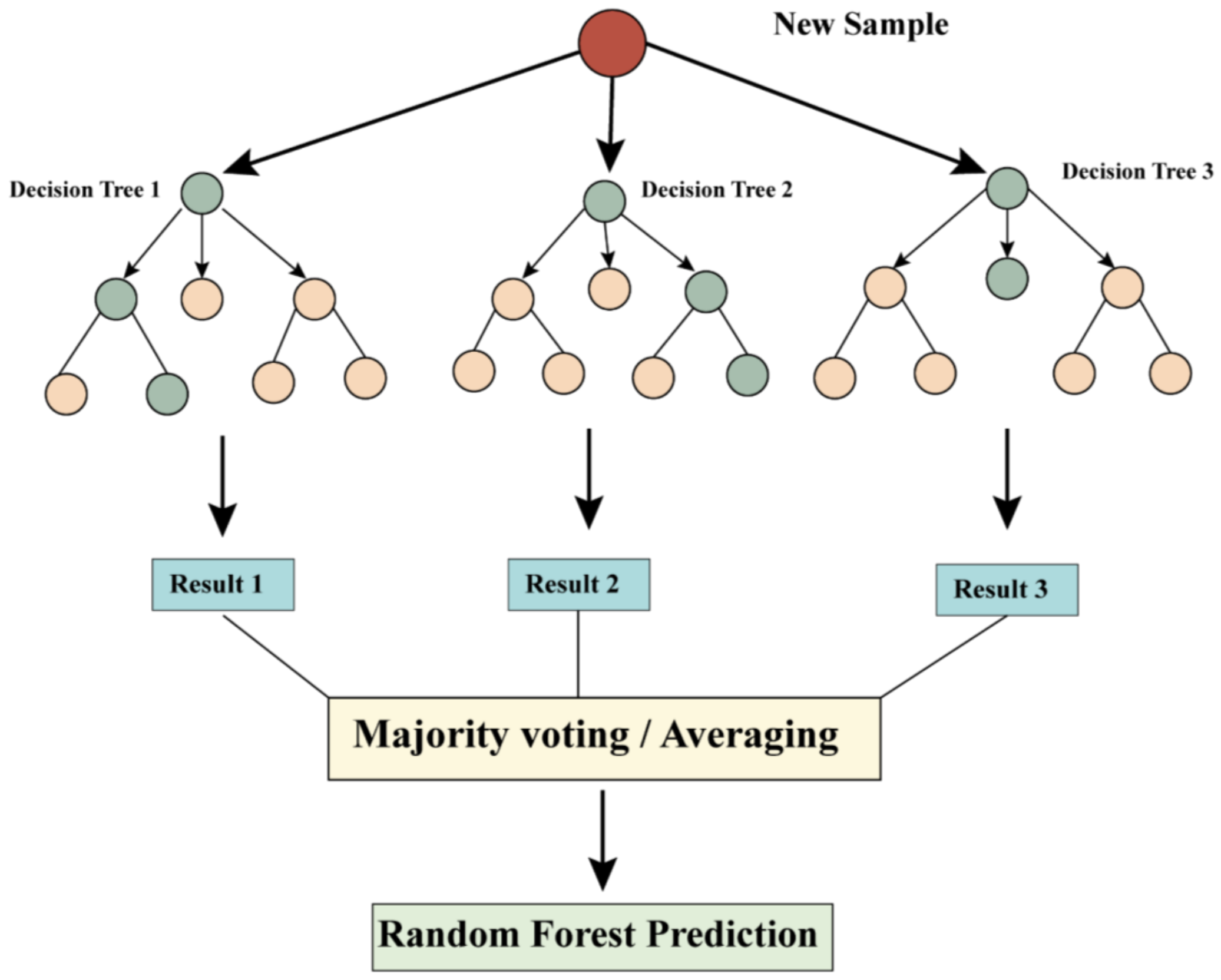

**Figure 7.** Random Forest block diagram.

## 5. Results and Discussions

In this section, our extracted findings have been organized into five subsections, representing the sample data size, measured body parameters, performance metrics, AI models for sleep staging and sleep disorder detection, and the summary. We answer the research questions to find reliable data sources and the size of datasets for sleep studies, the most utilized body parameters for sleep studies, the most widely used AI models, and the widely used metrics to evaluate the AI models.

## 5.1 Sample Data Size

The performance of the executed model is explicitly dependent on the datasets and number of samples used to train the AI models. Based on the training data and its extracted features, the classification of sleep stages and estimation of disorders can be made with a higher degree of accuracy.

**Table 2: Sample size and different datasets used in sleep stage classification.**

| Reference | Dataset used | Sample size |
|---|---|---|
| [24] | Sleep-EDFX | 77 subjects |
| [25] | Sleep-EDF | 42308 epochs |
| [26] | Sleep-EDFX, Sleep Heart Health Study (SHHS), and Haaglanden Medisch Centrum Sleep-Staging Dataset | 6, 441 subjects (SHHS), 348 whole night recordings (Sleep-EDFX, HMCS) |
| [27] | Sleep-EDF and MIT-BIH polysomnographic Dataset | 98 subjects |
| [28] | Sleep-EDF and MASS-SS3 | 82 subjects |
| [29] | The Nationwide Children's Hospital (NCH) | 3984 subjects |
| [30] | Sleep-EDF | 20 subjects |
| [3] | PhysioBank | 153 EEG recordings |
| [31] | Soonchunhyang University Bucheon Hospital | 602 subjects |
| [32] | Sleep-EDFX-8, Sleep-EDFX-20, Sleep-EDFX-78, and SHHS | 435 subjects |
| [33] | Proprietary data | 112 subjects |
| [34] | SHHS | 2274 subjects |
| [35] | Sleep-EDFX and University College Dublin Sleep Apnea Database (UCDDB) | 45 subjects |
| [36] | University of British Columbia Hospital | 12 subjects |
| [37] | DREAMS database from the University of MONS–TCTS Laboratory and Université Libre de Bruxelles - CHU de Charleroi Sleep Laboratory | 20 subjects |
| [38] | Sleep-EDF and Sleep-EDFX | 142700 samples |
| [39] | SHHS | 1000 registers |
| [40] | Proprietary data | 184 subjects |
| [41] | Sleep-EDFX | 20 subjects |
| [42] | SHHS, clinical sleep laboratories at the Charité Hospital Berlin (CHB) | - |
| [10] | Sleep-EDF (2013) | 30 subjects |
| [43] | Sleep-EDF (2013 & 2018) | 98 subjects |
| [44] | Sleep-EDF and CinC2018 Datasets | 697 subjects |
| [12] | Proprietary data | 106 subjects |
| [45] | Physionet CAP Sleep Dataset 2012 | 108 subjects |
| [46] | MESA sleep study | 43 subjects |
| [7] | Sleep-EDF (2013 & 2018) | 258 PSG recordings |
| [47] | HMC, SHHS, UCD Apnea Dataset, Telemetry, DREAMS, ISRUC | 443 PSG recordings |
| [48] | MESA, MrOS, and Apple Watch Datasets | 1456 subjects |
| [49] | Sleep-EDFX, St. Vincent University Hospital, and UCD | 125 subjects |
| [50] | Proprietary data | 28 subjects |
| [51] | NSRR by WSC, CAP, Sleep EDF, ISRUC, MIT-BIH, and St. Vincent University Hospital sleep Dataset | 2431 subjects |
| [52] | Proprietary data | 2889 subjects |
| [53] | Nocturnal sleep sound Dataset | 700 audio signals and 140 recordings |
| [54] | Proprietary data | 111 subjects |
| [55] | SHHS, WSC, MESA, MNC, NCHSDB, SNUH and CCC Datasets | 11555 recordings for training and validation |
| [56] | SIESTA, Somnolyzer, SOMNIA, and HealthBed Datasets | 1171 subjects |
| [57] | Cleveland Clinic Sleep Registry | 72 subjects |
| [58] | Sleep-EDFX | 153 subjects |
| [59] | Sleep-EDF, Sleep-EDFX, and DREAMS Datasets | 24 subjects (Sleep EDF), 20 nights of PSG recordings (DREAMS Subject databases) |
| [8] | SIESTA, and Eindhoven Dataset | 352 subjects |
| [60] | SHHS, MESA, and CinC Dataset | 993 subjects |
| [61] | SIESTA Dataset | 292 subjects |
| [9] | One night sleep recordings from 3 studies | 158 participants (training and testing) |
| [11] | 21 survey Datasets | 15660 subjects |
| [62] | Sleep-EDF Dataset | - |
| [63] | Sleep-EDFx, ISRUC-Sleep | 178 subjects |
| [64] | Sleep EDF, DREAMS subject Dataset | 48 subjects |
| [65] | Sleep EDF, Sleep EDFx | 69 subjects |
| [66] | Sleep EDF, Sleep EDFx | > 40 subjects |

The databases enlisted in Table 2 for classifying the sleep staging problem are mostly from the Physionet Sleep-EDF repositories, and a few are from proprietary data collected by the authors. Researchers have fused up to 7 different datasets in some studies for diversity [55]. The enrolled human subjects are considered for the sample size, but if no subjects are mentioned, then the PSG or other recorded signals or sample numbers are mentioned as the sample size. The range of subjects even spans from a few to several thousand. For instance, Casciola et al. [36] analyzed data from only 12 hospital-enrolled subjects, and Perslev et al. [11] included 15660 subjects in their study from 21 datasets. For inclusion, in most of the articles with different datasets and different samples, the summation of the subjects or any other tracks (e.g., audio signals, epochs, and samples) is considered. The number of datasets collected from hospitals or universities, including University College Dublin and St. Vincent University Hospitals, is also notable.

**Table 3: Sample size and different datasets used in sleep disorder detection.**

| Reference | Dataset used | Sample size |
|---|---|---|
| [67] | Sleep-EDF | 197 PSG recordings |
| [4] | CAP sleep and TURIN Dataset | 56 subjects |
| [68] | CAP sleep Dataset | 77 subjects |
| [69] | Lab-PSG Dataset | 1579 subjects |
| [14] | Stanford Technical Analysis and Sleep Genome Study (STAGES) | 80 subjects |
| [5] | PhysioNet Apnea-ECG Dataset | 70 PSG recordings |
| [70] | Enrolled HSAT patients | 14 subjects |
| [71] | Enrolled hospital patients | 3972 data samples |
| [72] | Primary data | 40 subjects |
| [73] | Taiwan National Health Insurance Research Dataset(NHIRD) | 14788 subjects |
| [74] | Proprietary data | 32 subjects |
| [75] | Proprietary data | 83 subjects |
| [76] | CAP sleep Dataset | 30720 samples |
| [77] | National Health and Nutrition Examination Survey (NHANES 2017–2020) | 4055 subjects |
| [78] | Proprietary data | 88 subjects |
| [46] | SHHS, UCD Sleep Apnea, and MIT-BIH Polysomnographic Dataset | 2691 subjects |
| [79] | PhysioNet Apnea-ECG Dataset | 70 subjects |
| [80] | Proprietary data | 78 subjects |
| [81] | UCD sleep apnea Dataset | 15 subjects |
| [82] | Proprietary data | 653 subjects |
| [83] | Proprietary data | 124 subjects |
| [84] | SHHS 1, WSC Dataset | 5243 subjects |
| [85] | Enrolled patients at the Northwell Health Sleep Disorders Center | 50 subjects |
| [86] | Enrolled patients at the clinic of Samsung Medical Center | 4014 subjects |
| [87] | Examined patients at the Samsung Medical Center | 1241 subjects |
| [88] | Visited patients at the Seoul National University Hospital | 98 subjects |
| [89] | Recruited patients at Seoul National University Hospital | 40 subjects |
| [90] | Patients enrolled in Tianjin Chest Hospital | 30 subjects |
| [91] | TMUH, SHH Datasets | 6399 subjects |
| [92] | Proprietary data | 50 subjects |

For sleep disorder detection-related research, Table 3 includes mainly proprietary datasets along with datasets from different institutes, whereas Table 2 includes a handful. The included subjects range from a few to several thousand. Enrolled hospital patients in the Samsung Medical Center, Seoul National University Hospitals, and the National Health Survey are some of the institutions that collect the databases used for research.

## 5.2 Body Parameters

Each generated body part's signals are significant in analyzing the nature of any usual distinctive work control pattern. Different characteristics can be obtained based on the measured body parameters, which are further used to classify different sleep stages and predict sleep-related disorders. According to our findings, different measurement types have been enlisted in Tables 4 and 5. Brain waves are the most common sources of signals, followed by the other PSG data, along with different physical symptoms. These frequently used body-generated signals exhibit meaningful potential features to be extracted. The percentage evaluation is done within each table count, where brain signals solely held the weights of 36% for sleep stage identification and 22.22% for sleep disorder detection.

**Table 4: Different body parameters used in sleep stage classification.**

| Reference | Measurement of the body parameters | Percentage of use |
|---|---|---|
| [24, 25, 28, 29, 3, 32, 37, 41, 43, 44, 9, 93, 49, 51, 57, 58, 64, 65] | Brain waves | 36% |
| [27, 34, 36, 39, 47, 8] | Eye movement, brain activity, heart rate, and muscle movement | 12% |
| [30, 31, 35, 50, 59, 62] | Eye movement, brain activity, and muscle movement | 12% |
| [34, 45, 52] | Eye movement, brain activity, heart rate, muscle activity, and respiratory flow | 6% |
| [33, 60, 61] | Heart rate | 6% |
| [38, 10] | Brain waves and eye movement | 4% |
| [55, 56] | Heart rate and body movement | 4% |
| [63] | Brain waves, eye movement, muscle activity, and respiration | 2% |
| [66] | Brain waves, eye movement, muscle activity, nasal respiration, and temperature | 2% |
| [40] | Brain wave, eye movement, heart rate, and breathing (flow) | 2% |
| [36] | Eye movement, brain activity, heart rate, muscle movement, respiratory flow, pulse oximetry, audio-equipped video | 2% |
| [42] | Eye movement, brain activity, heart rate, muscle movement, respiratory flow, oxygen saturation, and blood circulation volumetric change | 2% |
| [12] | Temperature, heart rate, acceleration, and circadian rhythm | 2% |
| [48] | Wrist accelerometry and heart rate | 2% |
| [53] | Coughing, laughing, screaming, sneezing, snoring, sniffling, and farting | 2% |
| [54] | Body movements, sleep bruxism, and snoring | 2% |
| [46] | Body movement | 2% |

**Table 5: Different body parameter signals used in sleep disorder detection.**

| Reference | Measurement of the body parameters | Percentage of Use |
|---|---|---|
| [67, 68, 76, 78, 13, 90] | Brain waves | 22.22% |
| [14, 5, 79] | Heart rate | 11.11% |
| [70, 81, 89] | Respiratory flow/level | 11.11% |
| [6, 4, 86] | Brain wave, heart rate, muscle movement, and eye movement | 11.11% |
| [77, 91] | Anthropometric biomarkers | 7.4% |
| [69, 83] | Nasal flow, oxygen saturation, and brain wave | 7.4% |
| [4] | Muscle activity, eye movement, and heart rate | 3.7% |
| [71] | Nasal pressure, oxygen desaturation level, brain wave, and heart rate | 3.7% |
| [72] | Diffusion tensor imaging | 3.7% |
| [75] | Heart rate and blood flow volumetric change | 3.7% |
| [80] | SpO2 and Changing heart rate | 3.7% |
| [85] | Airflow, SpO2, and Rib movements | 3.7% |
| [88] | Cortical activities | 3.7% |
| [92] | Snore, oxygen saturation, arousal level, and other body signals | 3.7% |

### 5.3 Type of Metrics

Determining whether a machine learning model is suitable for solving a specific problem is a crucial concern, and performance metrics are used to evaluate it. Metrics such as accuracy, F1 score, sensitivity, and specificity are commonly used for classification problems in machine learning algorithms. In contrast, Mean Absolute Error (MAE), Mean Squared Error (MSE), Root Mean Squared Error (RMSE), etc., are widely used for regression problems. Among these metrics, accuracy or precision is the most frequently used for sleep studies. In fact, accuracy was included in 86.42% of papers alongside multiple other evaluation metrics. However, accuracy may not be optimal if the data distribution is unevenly balanced. It is most commonly used in sleep studies due to its straightforward working principle. The accuracy can be denoted as:

$$Accuracy = \frac{T_P + T_N}{T_P + T_N + F_P + F_N}$$

Here, $T_P$, $T_N$, $F_P$, and $F_N$ denote the true positive, true negative, false positive, and false negative, respectively. For accuracy, the true negative values are also incorporated. For example, if there are any missing sleep stage categorization values, they are also included. This is being updated in calculating precision.

$$Precision = \frac{T_P}{T_P + F_P}$$

But, for the sake of simplicity, most of the metrics with accuracy were extracted, and a few with precision performance were included. Cohen's Kappa is another index that considers the random probable chances, leading to less inaccuracy. Its structural equation is denoted as:

$$Kappa = \frac{real\ accuracy - expected\ accuracy}{1 - expected\ accuracy}$$

Kappa is used in 39.51% of papers, along with the other metrics. It's the third most widely used metric in this review. The recall or sensitivity is similar to the precision, but the actual positive values are tried to be measured by mistakenly identified negative wrong values, which can be expressed as:

$$Sensitivity = \frac{T_P}{T_P + F_N}$$

Specificity is also a commonly used metric like sensitivity, evaluated based on a true negative.

$$Specificity = \frac{T_N}{T_N + F_N}$$

Based on the recall and precision values, the evaluation metrics F1 Score are obtained when both parameters are important to evaluate; it adds more significance. Our findings show that the F1 score is the second most frequently used performance metric, at 46.91%. The formula which calculates-

$$F1\ Score = 2 * \frac{Precision\ *\ Recall}{Prcision\ +\ Recall}$$

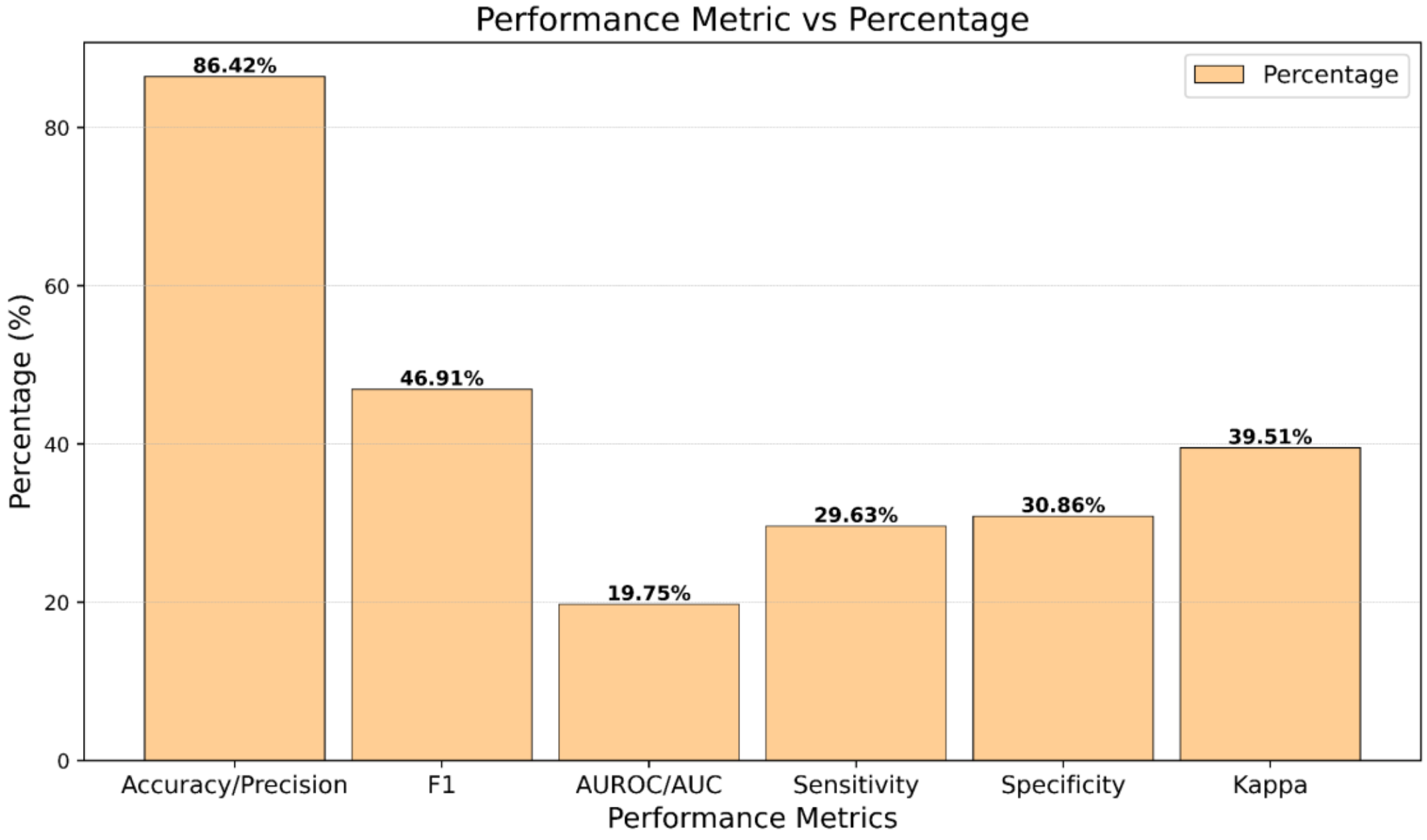

**Figure 8.** Different types of performance metrics are used to measure AI models.

### 5.4 AI Models for Sleep Stage Classification and Sleep Disorder Detection

Tables 6 and 7 exhibit the individual paper's proposed AI models and the degree of the best-performing model with performance metrics and its evaluation methods. Among the 81 papers, 52 have been enlisted as sleep-scoring related, and the remaining 29 articles have been segmented into sleep disorder-oriented works. In our study, 13 different AI parent algorithms were found. Within the parent AI methods, we have 32 unique subcategories of AI models. These various approaches reflect the diversified interest in this field of research. Figure 9 shows a clear idea about the parents and their corresponding children or the hybrid models used in different studies. NN has maximum modified child models. This study also found that CNN, RNN, AE, ResNet, TCN, DNN,

MRCNN, PPG-NN, BNN, Sle-CNN, and Fuzzy-NN are used as the children of neural networks. EL includes BO-ERTC, EBT, EBoot, LightGBM, XGBoost, CATBoost, Bagging. SVM includes CSVM, SVM-RBF, and SVM, which are used independently. GRU includes BiGRU and GRU independently. LSTM includes LSTM's main architecture and a hybrid named BiLSTM. DT, Fuzzy Network, HAC HHM, K-means, KNN, LASSO, NB, and RF architectures are the main and child architectures. In Table 6, along with the AI models and the metrics, the number of sleep stages according to some sleep scoring standards like the AASM, Rechtschaffen and Kales (R&K), and other criteria are included.

**Table 6: AI models for sleep stage classification**

| Reference | Proposed AI models | Best fitted model | Evaluation methods | Performance metrics | No. of Sleep Stages | Standard for sleep scoring |
|---|---|---|---|---|---|---|
| [24] | ANN+CNN | ANN+CNN | Skilled medical technicians | Accuracy 96%, F1 96.49% (wake) | 7 | R&K, AASM |
| [25] | NAMRTNet with TCN | ResNet +TCN | 20-fold CV | Accuracy 86.2%, F1 79.8%, Kappa 0.81 | 5 | AASM |
| [26] | BiRNN + GRU | BiRNN + GRU | 30-fold CV | Accuracy 87.3%, F1 82.50%, | 5 | R&K |
| [27] | Bi-LSTM + CNN | BiLSTM + CNN | 10-fold CV | Accuracy 90.83% (Avg.), F1 87.05% (Avg.), Kappa 0.85 | 5 | R&k |
| [28] | Siamese CNN, Siamese AE | Siamese CNN, Siamese AE | 10-fold CV | Accuracy 86.2%, F1 80.2%, Kappa 0.8 | 5 | AASM, R&K |
| [29] | CNN+BiLSTM | CNN + BiLSTM | 77% training and 33% testing | F1 90% | 5 | AASM |
| [30] | T-MCCFNN | T-MCCFNN | 70% training and 30% testing | Accuracy 85.3%, Precision 87.3%, F1 85.3% | 6 | R&K |
| [3] | ADNN | ADNN | 10-fold CV | Accuracy 98.75 | 6 | R&K |
| [31] | CNN+RNN | CNN + RNN | 482 datasets for training, 48 for validation, and 72 for testing | Kappa 0.84 | 5 | Helsinki Declaration, AASM |
| [32] | CNN+Bi-LSTM | CNN + BiLSTM | 20-fold, and LOSO CV | Accuracy 87.28%, F1 54.97%, Kappa 0.82 (Avg.) | 5 | R&K |
| [33] | CNN+GRU | CNN + GRU | 89 subjects for training and 23 subjects for testing | Accuracy 86.4% | 5, 3 | AASM |
| [34] | CNN | CNN | 70% training, and 30% for test and validation | Accuracy 91.45%, F1 89%, Kappa 0.84 | 3 | AASM |
| [35] | CNN | CNN | One subject for testing and one subject for validating | Accuracy 82.6%, F1 80%, Kappa 0.77 (Avg.) | 5 | R&K, AASM |
| [36] | CNN + LSTM | CNN + LSTM | Leave-one-out CV | Accuracy 75.5% (Avg.) | 5 | AASM |
| [37] | ResNet | ResNet | 4-fold CV | Accuracy 87.8% | 5 | AASM |
| [38] | CNN | CNN | 70% for training and 30% for test and validation | Accuracy 98.06%, F1 97%, Sensitivity 98% (Highest) | 6 | R&K |
| [39] | CNN | CNN | Three different datasets for training, testing, and validation | Precision 78%, F1 76%, Sensitivity 75%, Kappa 0.83 | 5 | AASM |
| [40] | BNN, K-NN, DT | BNN | Leave-one-out and 10-fold CV | Accuracy 89.15% (Avg.) | 5 | - |
| [41] | SVM, DT, NN, KNN, NB, LDA | DT | - | Accuracy 93.13%, Sensitivity 89.06%, Specificity 98.61% | 6 | R&K |
| [42] | TCN | TCN | 5-fold | Accuracy 73.8%, Kappa 0.59 | 4 | AASM |
| [10] | RF, KNN, SVM, CNN+LSTM | CNN + LSTM | 80% for training and 20% for validation | Accuracy 87.4%, F1 44.6%, Specificity 87% (Avg.) | 5 | - |
| [43] | MLTCN_TCN, MLTCN_LSTM | MLTCN_TCN | Leave-One-Subject-Out and 20-fold CV | Accuracy 82.6%, Kappa 0.76 (Avg.) | 5 | AASM |
| [44] | MRCNN | MRCNN | 5-fold CV, Independent subject CV | Accuracy 92.6%, Kappa 0.736 (5-fold CV) | 5, 6 | AASM |
| [12] | Light GBM | LightGBM | 5-fold CV. | Accuracy 96%, F1 78%, Sensitivity 80.74%, Specificity 98.15% (full model- wake) | 2, 4 | AASM |
| [45] | CSVM with L-Tetrolet pattern | CSVM with L-Tetrolet pattern | 10-fold CV | Accuracy 92.93%, F1 92.57%, Sensitivity 87.84%, Specitivity 98.02% (Avg.) | 6 | - |
| [46] | HMM | HMM | - | Accuracy 85.7%, Sensitivity 99.30%, Specificity 36.4%, Kappa 0.446 | 2 | - |
| [7] | RNN+CNN | RNN + CNN | K-fold CV (20 & 10) | Accuracy 84.26%, F1 79.66%, Kappa 0.79 | 5 | R&K |
| [47] | CNN+LSTM | CNN + LSTM | 80% for training, 20% for validation | Kappa 0.675 (Avg.) | 5 | R&K, AASM |
| [48] | CNN+LSTM | CNN + LSTM | 75% for training, 12.5% for validation, and 12.5% for independent subset and 5-fold CV | Accuracy 79%, F1 80%, Sensitivity 77%, Specificity 89%. | 3, 4 | AASM |
| [94] | MAResnet-BiGRU, Resnet-BiGRU, CNN- | MAResnet-BiGRU | 10-fold CV | Accuracy 81.73% (Avg), Kappa 0.789. | 5 | R & K |
| | BiGRU, MAResnet, CNN-GRU, and ResNet | | | | | |
| [49] | GRU (RNN) | GRU (RNN) | 10-fold CV (Sleep-EDFx) 5-fold CV (UCD) | AUROC 93.95%<br>Kappa 0.7095<br>(Avg- Sleep-EDFx, UCD) | 5 | R & K |
| [50] | RF | RF | 70% for training, 20% for validation, and 10% for test | Accuracy 93.9%, F1 82.8%, Kappa 0.745. | 5 | AASM |
| [51] | SVM, EBT, DT, KNN, and NB | EBT | 10-fold CV | Accuracy 81.5%, Sensitivity 81.3%, Specificity 95.3%, Kappa 0.702 (CT-12). | 5 | R&K, AASM |
| [52] | Sle-CNN | Sle-CNN | 83% for training and 17% validation | Accuracy 89.6% | 5 | AASM |
| [53] | ANN, Support Vector Classifier, KNN, NB, and CNN | ANN | 5-fold CV | Accuracy 97.4%, F1 97% | 5 | - |
| [54] | VDANN +LSTM | VDANN+LSTM | 5-fold CV | Precision 84.8%, F1 83%. | - | - |
| [55] | DNN+LSTM | DNN+BiLSTM | 80% for training and 20% for validation | Accuracy 78.8%, Kappa 0.64<br>(Avg- CCC, SNUH). | 4 | R&K, AASM |
| [56] | PPG-NN | PPG-NN | 1113 recordings for training and 394 recordings for validation | Accuracy 77.8%, Kappa 0.638 | 4 | R&K, AASM |
| [57] | K-NN | K-NN | 5-fold CV | Accuracy 80.2%, Specificity 85.3% | 3 | AASM |
| [58] | CNN, Bi LSTM, and EDA with a multi-head attention mechanism | CNN+multi-head attention+ BiLSTM | 10-fold CV | Accuracy 90.79%, F1 86.73%,<br>Kappa 0.87 | 5 | R&K |
| [59] | RF, K-NN, Bagging, and AdaBoost | Bootstrap aggregating (bagging) | 70% for the training and 30% for test | Sensitivity<br>99.29% (AWA) 45.85% (S1) 85.8% (REM) | 2, 3, 4, 5, 6 | R&K, AASM |
| [8] | LSTM | LSTM | 4-fold CV | Accuracy 76.36%, F1 81% (REM), Sensitivity 83.69% (REM), Kappa 0.65. | 4 | AASM, R&K |
| [60] | CNN | CNN | - | Accuracy 76.33%, Kappa 0.63 (Avg) | 4 | AASM |
| [61] | LSTM | LSTM | 4-fold CV | Accuracy 77%, Kappa 0.61 | 4 | R&K |
| [9] | RF | RF | 5-fold CV | Accuracy 93.31%, Kappa 0.85 (Nonwear, test) | 4 | |
| [11] | CNN+RF | CNN+RF | Leave-one-out CV | F1 79% (Overall) | 5 | AASM |
| [62] | NCA, SVM, ANN | ANN | 70% for training, | Accuracy 90.33% | 5 | R&K |
| | | | 30% for test and validation | Sensitivity 90.57%<br>Specificity 97.58% | | |
| [63] | ResNet+LSTM | ResNet + LSTM | - | Accuracy 84.18%<br>F1 80.95% (Avg) | 5 | AASM |
| [67] | BiRNN | BiRNN | 15-fold CV | Accuracy 70-85%, F1 80%, Specificity 95% | 5 | R&K |
| [64] | Linear Discriminant, Bagged Trees, SVM, Boosted Trees, KNN, RUSBoosted Trees | Bagged Trees | 10-fold C +V | Accuracy varies from 98.6% to 91.8% for Sleep-EDF and from 94.9% to 78.2% for Dreams Subjects | 6 | R&K |
| [65] | Bagging | Bagging | - | Avg. Accuracy 93.79% (Sleep- EDF)<br>Avg. Accuracy 91.69% (Sleep- EDFX) | 6 | R&K, AASM |
| [66] | SVM | SVM | 10-fold CV | Accuracy 89.8%, & 91.4%<br>Sensitivity 71.4%, & 67.6%<br>Specificity 97.5%, & 97.6%<br>(For Sleep-EDF, & Sleep-EDFx- 6 class) | 6 | R&K, AASM |

**Table 7: AI models for sleep disorder detection**

| Reference | Disorders | Proposed AI models | Best fitted model | Evaluation methods | Performance metrics |
|---|---|---|---|---|---|
| [4] | RSWA | K-NN, SVM | K-NN | 5-fold CV | Accuracy 86.96%, Sensitivity 93%, Specificity 75%. |
| [68] | Insomnia, Nocturnal frontal lobe epilepsy (NFLE), Narcolepsy, RBD, Periodic leg movement disorder (PLM), and sleep-disordered breathing (SDB) | EBT, EBooT, SVM, KNN | EBT, EBoot | 10-fold CV | Accuracy 98.25% (Avg.) |
| [69] | Insomnia disorder (ID) and OSA with respiratory arousal threshold (ArTH) phenotypes | LR, K-NN, NB, RF, and SVM | RF | 10-fold CV | Accuracy 80.06%, F1 80.41%, AUC 93.61% |
| [14] | Major depressive disorder (MDD) | BO-ERTC and SVM | BO-ERTC | 10-fold CV | Accuracy 86.32%, F1 86%, AUC 94.8%, Sensitivity 85.85%, Specificity 86.49% |
| [5] | OSA | CNN+LSTM | CNN+LSTM | 35 PSG recordings for training and 35 recordings for | Accuracy 96.1%, Sensitivity 96.1%, Specificity 96.2%, Kappa 0.92 |
| | | | | test | |
| [70] | Sleep apnea | LSTM | LSTM | Leave-One-Subject-Out (LOSO) cross-validation | Kappa 0.81 (Perfect agreement) |
| [71] | Sleep apnea | CNN, LDA, SVM, BRT, and ANN | CNN | Training data 20% and test data 80% | Accuracy 90.43%, F1 90.98%, Sensitivity 93.21%, Specificity 90.37% (Avg.) |
| [72] | REM, RBD | SVM | SVM | Training on 32 subjects and testing on eight subjects | Accuracy 87.5%, AUC 90%, Sensitivity 80%, Specificity 100% |
| [73] | Sleep disorder in an asthma cohort | KNN, SVM, RF, RNN, LSTM, GRU, and CNN | CNN | Two-image datasets validation | Accuracy 92.3%, F1 91.3%, AUC 93.4%, Sensitivity 89.3%, Specificity 96.6% |
| [6] | iRBD | LR, RF, and XGBoost | RF | Leave-One-Out Cross-Validation (LOO-CV) | Accuracy 94%, AUC 95%, Sensitivity 95%, Specificity 92% |
| [75] | Sleep deprivation, cold hands and feet, and the Shanghuo syndrome | Ensemble learner, Kernel classifier, KNN, SVM, NB, NN, and DT | Ensemble learner, kernel classifier, and KNN | - | Accuracy 84%, AUC 84% (Avg.) |
| [76] | Bruxism, narcolepsy, sleeplessness, irregular leg movements, fast eye movement behavior, and breathing abnormalities | Adaboost, RF, KNN, and SVM-RBF | SVM-RBF | 70% of features for training 30% for testing | Accuracy 93.5%, F1 66.9% |
| [77] | Insomnia | XGBoost, RF, ADABoost, and ANN | XGBoost | 10-fold CV | AUROC 87%, Sensitivity 77%, Specificity 77% |
| [78] | Severe chronic disorders of consciousness | Genetic algorithm and SVM | SVM | 77% for training and 33% for test | AUC 90% |
| [46] | Sleep apnea | CNN | CNN | 10-fold CV | Accuracy 69.9%, Kappa 0.7 (SHHS) |
| [79] | OSA | CNN | CNN | 10-fold CV | Accuracy 92.4%, F1 90.6%, Specificity 92.6% |
| [80] | OSA | NN | NN | Internally validated by 1000 bootstrapping samples | Accuracy 91%, AUROC 96.1%, Sensitivity 71.4%, Specificity 95.3%, Kapap 0.687. |
| [81] | Apnea and hypopnea | Fuzzy network | Fuzzy network | - | Accuracy 97.5% (detecting apnea) 95.2% (detecting hypopnea) |
| [82] | OSA | CATBOOST, | CATBOOST | 10-fold CV | Accuracy 71%, F1 |
| | | LIGHTGBM, RBFSVM, ET, LR | | | 73%, AUC 76%, Sensitivity 75%, Specificity 66% |
| [83] | Sleep-disordered breathing | SVM, RF, and LR | SVM, RF, LR | Leave one subject out (LOSO) CV | Accuracy 80% AUROC 94% |
| [84] | OSA | LR, SVC, K-NN, DT, ET, Adaboost, GNB, and RF | LR | 10-fold CV | AUROC 77%, Sensitivity 73%, Specificity 70% (Avg.) |
| [85] | OSA | CNN | CNN | Leave one subject out (LOSO) CV | Accuracy 84% |
| [86] | Obstructive sleep apnea syndrome(OSAS) | RF, XGBoost, LightGBM, CatBoost, K-means, Bisecting K-means algorithm, and GMM | HAC, CATBOOST, GMM, LightGBM, and K-means | 5-fold CV | Accuracy 87.52% (CatBoost, low OSAS) Accuracy 86.01% (LightGBM, moderate OSAS Accuracy 91.11% (LightGBM, severe OSAS) |
| [87] | Insomnia and excessive daytime sleepiness (EDS) | DT, SVM, ET, Gradient Boost, K-NN, RF, and CNN+ | CNN + Attention model | 80% for training and 20% for test | Accuracy 93%, F1 93% |
| [88] | iRBD | CNN | CNN | 80% for training and 20% for evaluation | Accuracy 99.81%, AUROC 99.41% (3D-CNN) |
| [89] | Insomnia | LASSO, PCA, and SVM | LASSO and SVM | Leave-one-out CV | Accuracy 80%, F1 79%, Specificity 81% |
| [90] | Sleep apnea syndrome (SAS) | RF, K-NN, SVM | RF | 10-fold CV | Accuracy 86%, F1 87%, Kappa 85% |
| [91] | OSA | SVM | SVM | Validated on the SHH dataset | Accuracy 81%, AUC 83.08%, Sensitivity 76.08%, Specificity 76.55% (Avg.) |
| [92] | Sleep apnea | DNN+RNN+GRU+LSTM | DNN+RNN+GRU+LSTM | N/A | Accuracy 95.76% F1 95.68% |

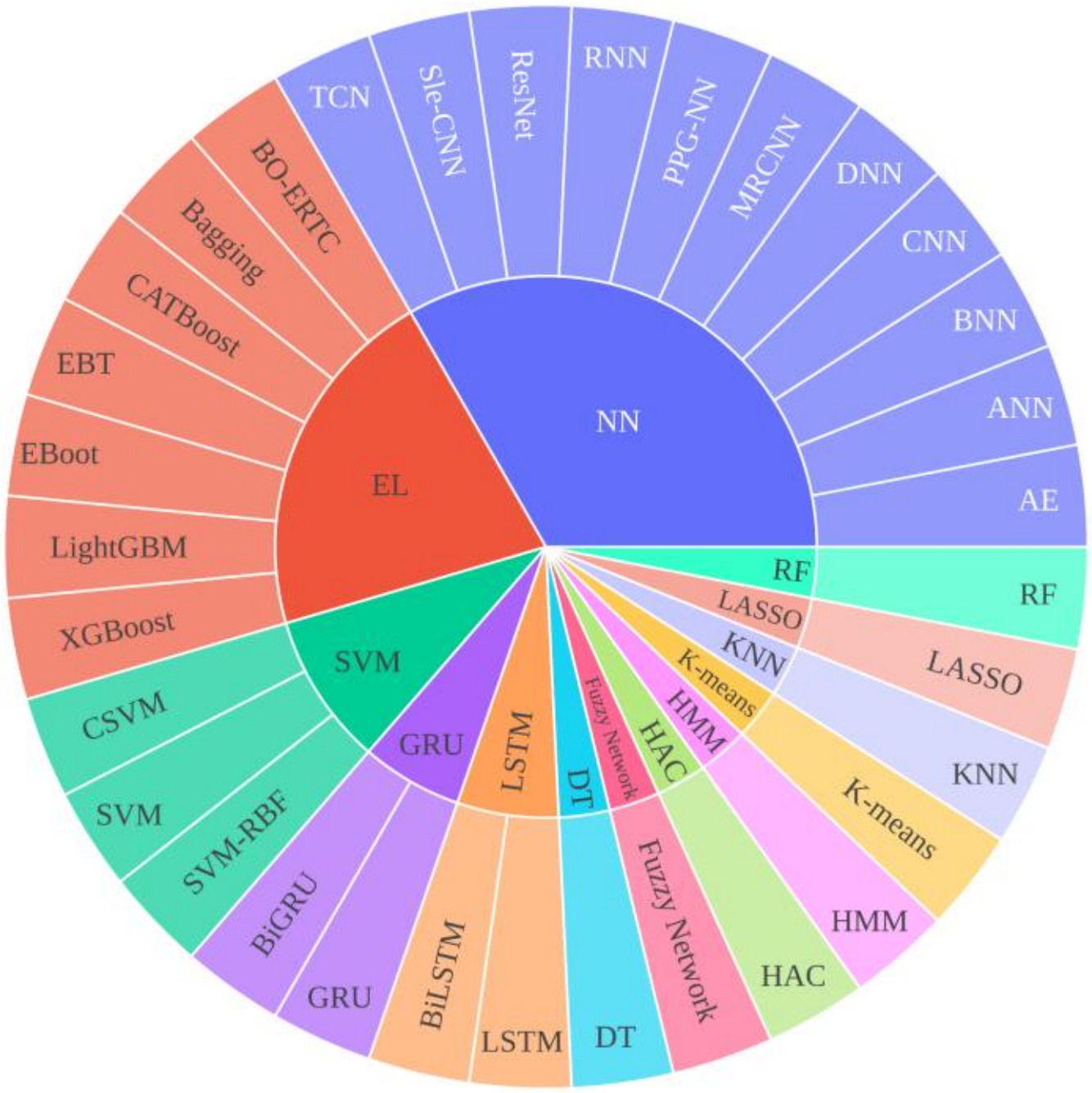

**Figure 9.** Unique parent and child AI models.

Figure 10 shows that the NN has the largest contribution, which is 47% of the total papers. The second largest contribution is from LSTM, which is 15% of the collected data. The third most common algorithm is EL, consisting of 12% of the parent algorithm of our collected data. SVM, RF, GRU, KNN, DT, and HMM have sequentially 7%, 6%, 5%, 2%, 1%, and 1% share each. The neural network has a variety of modified versions for different purposes. It is more popular and has a rich community as it can learn more complex patterns from data. Among these, SVM classifies finding the hyperplanes in the n-dimensional space based on the number of features. The incorporated kernel classifier can boost the required task performance of the SVM. SVM extracts non-linear information from the buffered EEG signals with higher accuracy [92]. On the other hand, ensemble learning is a multiclassifier that combines many learners and is used to improve predictive performance. A modified rotational SVM can extract even more complex features, which are rooted in the principle of ensemble learning [94]. For different applications, convolutional layers and kernels like RBF can be fused into SVM, building CSVM and SVM-RBF, depicted in Figure 10. Different parent and child architectures have their unique features. The features of the studied AI models are further explained in Table 8.

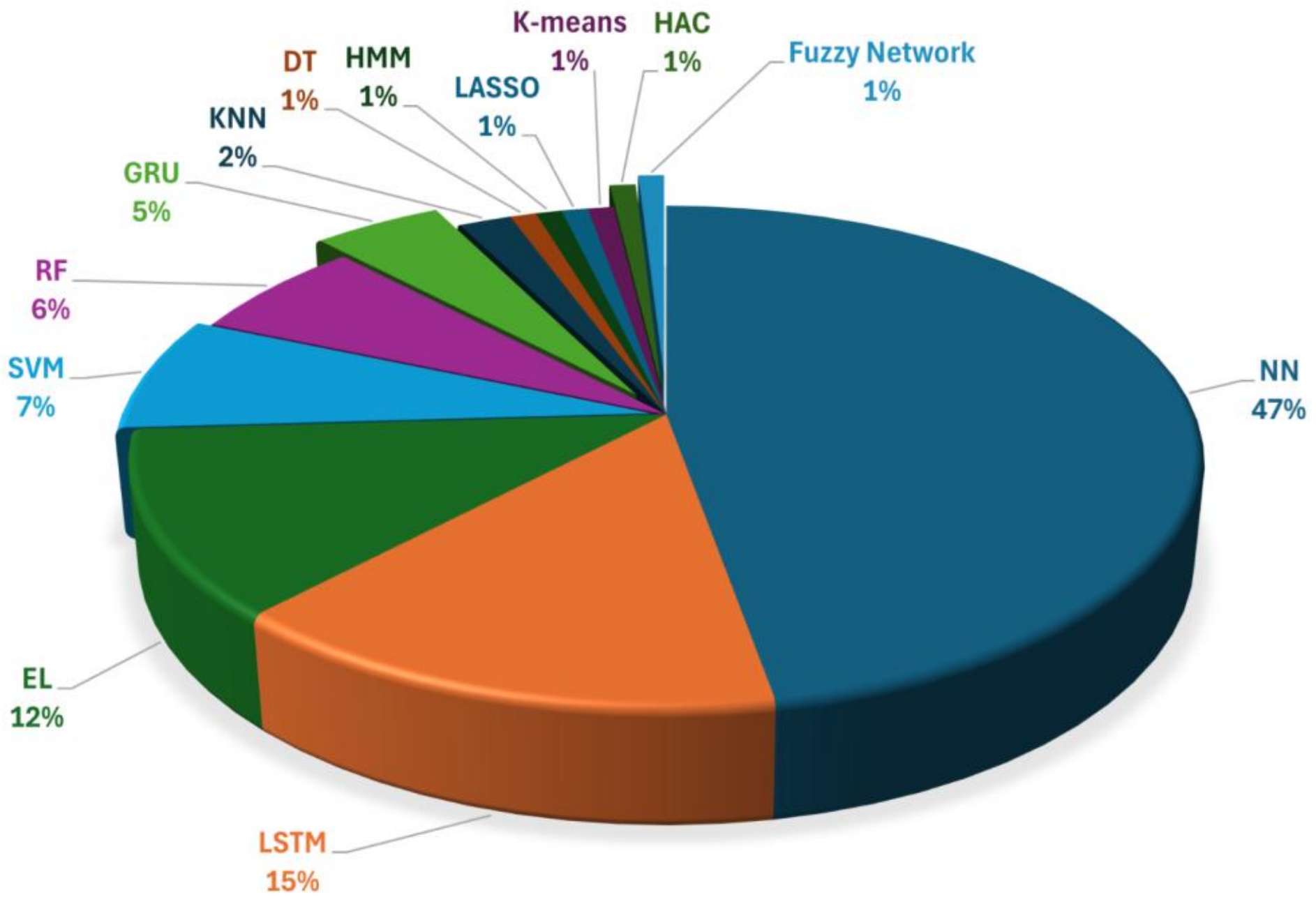

**Figure 10.** Frequency of parent models of the AI models.

**Table 8: Different child model's features and acronyms.**

| AI models | Full form | Main features |
|---|---|---|
| CNN | Convolutional Neural Network | It uses convolution as a form of shift-invariant feature extraction and learns through training to extract important features from time series data that are useful for classification [13]. |
| RNN | Recurrent Neural Network | Previous input data can be used in the network layer [7]. |
| AE | Autoencoder | The autoencoder uses convolutional layers in the encoder and transposed convolutional layers in the decoder. The encoder generates a latent representation (EN_L) of the input EEG, and the decoder reconstructs the input EEG with DE_L as the output [28]. |
| ResNet | Residual Network | Each layer is fed to the next layer. It also feeds to some layers that are a few hops away, helping to propagate larger gradients to the initial layer through backpropagation and training deeper networks [37]. |
| TCN | Temporal Convolutional Network | Can learn the dependencies between features of long-time series data [25]. |
| DNN | Deep Neural Network | Its deep architecture allows deep features to be learned by having more hidden layers [90]. |
| MRCNN | Multiscale Residual Convolutional Neural Network | Multiple residual blocks are incorporated with the CNN to extract features from various domains [44]. |
| PPG-NN | Photoplethysmographic Neural Network | A developed classifier that uses the IHR and activity counts for input and CNN for feature extraction [56]. |
| BNN | Bayesian Neural Network | A neural network with an output layer made of Bayes theorems for prediction [40]. |
| Sle-CNN | Sle-CNN | It is a CNN-based architecture fused with Kernel filters in each step [52]. |
| Fuzzy-NN | Fuzzy Neural Network | A fuzzy network with a convolutional kernel with an increased convergence training speed [30]. |
| LSTM | Long Short-Term Memory | LSTM is a form of recurrent neural network (RNN) architecture mainly designed to beat the issues of learning long-term dependencies in sequential data [93]. |
| BiLSTM | Bidirectional Long Short-Term Memory | Two layers of LSTM, onesies for the feedback from the output, will be used to merge with the previous data again. |
| SVM | Support Vector Machine | Separates classes by finding an optimal hyperplane [72]. |
| CSVM | Cubic Support Vector Machine | Cubic polynomial order with third-degree Kernel as a classifier algorithm [45]. |
| SVM-RBF | Support Vector Machine-Radial Basis Function | SVM classifier with the radial bias function [73]. |
| BO-ERTC | Bayesian optimised extremely randomized trees classifier | Decision tree-based classifier from the Gini split random datasets [14]. |
| EBT | Ensemble Bagged Trees | Average of the randomized different decision trees with relatively low variance and improving performance [65]. |
| EBoot | Ensemble Boosted Trees | A decision tree-based ensemble learning classifier combines weak and strong learners with higher accuracy [65]. |
| LightGBM | Light Gradient Boosting Machine | A gradient-boosting model using fewer memory deals with missing data and efficiently extracts features [12]. |
| XGBoost | eXtreme Gradient Boosting | Gradient boosting model with ranked covariates and features [6]. |
| CATBoost | Categorical Boosting | Average of important and high prioritized weights for the best stratification [80]. |
| Bagging | Bootstrap Aggregating | Independent learners lead to the highest performance on the mixed signals data [59]. |
| RF | Random Forest | RF, as ensemble learning, outperforms single tree-based models [57]. |
| GRU | Gated Recurrent Unit | GRU, a much simpler version of the LSTM, combines past and current data [93]. |
| BiGRU | Bidirectional Gated Recurrent Unit | BiGRU consists of two GRUs for forward and backward sequence processing. |
| KNN | K-Nearest Neighbors | classifies each element by taking the majority vote on the class of its K closest items [4]. |
| DT | Decision Tree | Creates hierarchy structures called nodes (leaf nodes and internal nodes) and constructs a model that can predict the label [72]. |
| HMM | Hidden Markov Model | An unsupervised model that classifies based on the activity counts from the Markov chain [46]. |
| NB | Naive Bayes | A probabilistic ML algorithm assumes the features are conditionally independent given the class label and utilizes probabilities to calculate the posterior probability of each class label given the features [72]. |
| LASSO | Least Absolute Shrinkage and Selection Operator | Reduce the risk of overfitting. It enables the proper number of features by limiting the number of effective features via regularization [87]. |
| K-means | K-means | Updating the model parameters around the centroid until it remains unchanged [86]. |
| HAC | Hierarchical | A clustering algorithm that nested and merged the individual and similar |
| | Agglomerative Clustering | clusters [86]. |
| Fuzzy Network | Fuzzy Network | A system based on the fuzzy logic theorem that deals with uncertain data [79]. |

## 5.5 Summary

This section extensively depicts the key findings as per the research questions. First, the number of datasets with samples from the extracted papers generalized the notion of getting the most information that impacts the AI model's performance. Most datasets are taken from the Physionet repository for the sleep stage classification and proprietary data, along with some institutional datasets for sleep disorder detection. In addition, the features based on the measured body parameters are likely to have a high objective-oriented relation and justify the performance of the distinctive AI model. The frequently found cases are the brain waves, with a maximum of 77% cases, along with other body parameters. Among the sleep stage and sleep disorder-related papers, the sole use of brain signals was 36% and 22.22%, respectively. The PSG data and physically noted symptoms also play a crucial role in sleep studies. The evaluation metrics for sleep studies are listed in Table 6 and Table 7, where we found that accuracy or precision is the most widely used metric, with the highest at 86.42%, along with other metrics. The F1 and Kappa are also extensively used metrics, with a share of 46.91% and 39.51%, respectively.

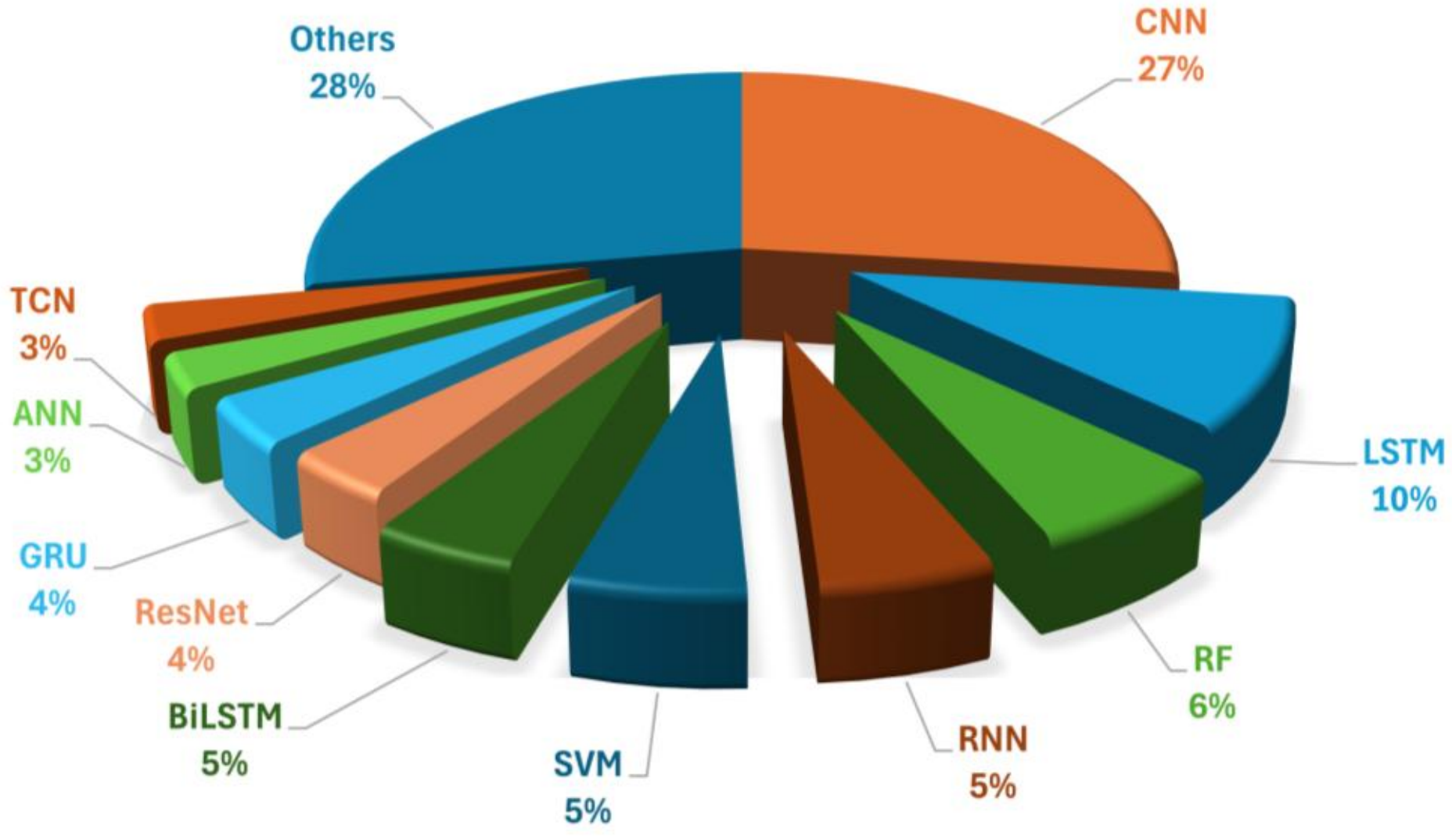

**Figure 11.** Top ten AI models.

Breaking parent nodal AI models, the top ten distinct single models are exhibited in Figure. 11. It includes only the child models used in the different research works; that's why those models have common parent and child names and a smaller portion in distinct singular representations. The top five frequently executed models are CNN, LSTM, RF, SVM, and RNN, with a share of 27%, 10%, 6%, 5%, and 5%, respectively. Since the brain's visual cortex influences CNN's working principle, it can learn from its raw, complex features. Adding multiple convolutional layers enables effective learning and categorizing features from the different signals. That's why CNN processes numerous body signals collected to determine sleep patterns with higher accuracy. There are many combinations or hybridizations with CNN, such as CNN+Bi-LSTM, CNN+GRU, CNN+LSTM, etc. As a result, it reflects that working with CNN to detect sleep stages and sleep disorders can be a good choice.

## 6. Future Research Directions

Despite significant advancements in automated sleep stage classification, a crucial gap remains in using sleep patterns to predict and diagnose sleep disorders. Current research has primarily focused on classifying sleep stages without utilizing these insights to identify underlying conditions such as sleep apnea, insomnia, or restless leg syndrome. To address this gap, future studies should prioritize predicting sleep disorders by integrating more comprehensive pattern analysis into existing models. One key area for improvement is enhancing the generalization of models trained on specific datasets. Future research could greatly benefit from integrating multimodal data. For example, combining EEG with other physiological signals, such as heart rate, respiratory patterns, or eye movements, may enhance the accuracy of both sleep stage classification and the diagnosis of sleep disorders. Additionally, incorporating explainable AI techniques presents a promising opportunity. By improving the transparency of AI models' decision-making processes, researchers can increase the reliability and trustworthiness of their predictions, which is crucial for medical professionals making critical clinical decisions.

In summary, future work should focus on the following areas:
1. Developing models to predict sleep disorders from sleep stage patterns.
2. Enhancing the generalization of models across diverse datasets.
3. Leveraging multimodal data for more accurate and comprehensive sleep analysis.
4. Incorporating explainable AI to improve the interpretability of model outputs for clinical use.

## 7. Conclusion

In this systematic literature review (SLR), we aimed to summarize recent advancements in sleep studies from 2016 to 2023, focusing on sleep stage classification and detecting sleep disorders. We addressed several key research questions, including the body parameters used for sleep analysis, the datasets and sizes employed in AI models, the most commonly applied AI models, and the evaluation metrics typically used to assess model performance. Our analysis revealed that brain signals, such as EEG, are used in nearly all studies, often in combination with other physiological parameters like heart rate and respiratory signals. The Physionet dataset emerged as the most widely used resource, alongside datasets from various universities and research centers. While most studies relied on publicly available datasets, a few researchers conducted experiments using their own data. Regarding AI models, neural networks, particularly CNN, RNN, and LSTM networks, were the most commonly employed. Traditional machine learning models, such as SVM and EF, were also frequently used. For model evaluation, accuracy and precision were the most popular metrics, given the classification nature of the problem, alongside other measures such as F1 score, sensitivity, and specificity. Overall, this review provides a comprehensive understanding of the current state of AI-driven sleep studies, highlighting key trends, common methodologies, and areas for future research.

The limitation of this study is the lack of financial support, which restricted access to paid research papers. As a result, we could not analyze certain relevant studies that required a subscription or purchase, which may have caused some potentially valuable insights to be missed. Despite this constraint, this study aimed to provide insight into sleep-related research by bringing a substantial number of freely available papers.

**Author contribution statement:**
All authors listed have contributed equally to this article's development and writing.

**Declaration of conflict of interest:**
The authors declare no conflict of interest.

**Funding information:**
This research received no specific grant from any funding agency.

**Data availability:**
Data will be made available on request.